\documentclass{article} % For LaTeX2e

\usepackage{iclr2021_conference,times}
\usepackage{amsmath,amsfonts,bm}

\def\eqref#1{equation~\ref{#1}}
\def\1{\bm{1}}

\DeclareMathAlphabet{\mathsfit}{\encodingdefault}{\sfdefault}{m}{sl}
\SetMathAlphabet{\mathsfit}{bold}{\encodingdefault}{\sfdefault}{bx}{n}

\DeclareMathOperator*{\argmax}{arg\,max}
\DeclareMathOperator*{\argmin}{arg\,min}

\usepackage{hyperref}
\usepackage{url}
\usepackage{booktabs} % for professional tables
\usepackage{multirow}
\usepackage{rotating, graphicx}%
\usepackage{caption}
\usepackage{subcaption}
\usepackage{wrapfig}

\makeatletter
\newcommand{\printfnsymbol}[1]{%
  \textsuperscript{\@fnsymbol{#1}}%
}
\makeatother

\title{Combining Label Propagation and Simple Models out-performs Graph Neural Networks}

\author{%
Qian Huang\textsuperscript{$\ddagger$}\thanks{Equal contribution}  , Horace He\thanks{Work done while at Cornell University}\hspace{0.1em} \printfnsymbol{1} , Abhay Singh\textsuperscript{$\ddagger$}, Ser-Nam Lim\textsuperscript{$\mathsection$}, Austin R.~Benson\textsuperscript{$\ddagger$}\\
Cornell University\textsuperscript{$\ddagger$}, Facebook\printfnsymbol{2}, Facebook AI\textsuperscript{$\mathsection$}\\
}

\newcommand{\framework}{C\&S}

\usepackage[capitalize,noabbrev]{cleveref}

\definecolor{mylinkcolor}{RGB}{0,0,140}
\hypersetup{colorlinks,allcolors=mylinkcolor,citecolor=mylinkcolor}

\newcommand{\xhdr}[1]{\vspace{0.0mm}\noindent{\textbf{#1.}}\hspace{0.5mm}}

\begin{document}

\maketitle

\begin{abstract}
Graph Neural Networks (GNNs) are the predominant technique for learning over graphs.
However, there is relatively little understanding of why GNNs are successful in practice and whether they are necessary for good performance.
Here, we show that for many standard transductive node classification benchmarks, we can exceed or match
the performance of state-of-the-art GNNs by combining shallow models that ignore the graph structure 
with two simple post-processing steps that exploit correlation in the label structure:
(i)  an ``error correlation'' that spreads residual errors in training data to correct errors in test data and
(ii) a ``prediction correlation'' that smooths the predictions on the test data.
We call this overall procedure Correct and Smooth (\framework{}),
and the post-processing steps are implemented via simple modifications to standard label propagation techniques from early graph-based semi-supervised learning methods.
Our approach exceeds or nearly matches the performance of state-of-the-art GNNs on a wide variety of benchmarks,
with just a small fraction of the parameters and orders of magnitude faster runtime.
For instance, we exceed the best known GNN performance on the OGB-Products dataset with 137 times fewer parameters
and greater than 100 times less training time.
The performance of our methods highlights how directly incorporating label information into the learning algorithm (as was done in traditional techniques)
yields easy and substantial performance gains.
We can also incorporate our techniques into big GNN models, providing modest gains.
Our code for the OGB results is at \url{https://github.com/CUAI/CorrectAndSmooth}.
\end{abstract}

\section{Introduction}

Following the success of neural networks in computer vision and natural language processing,
there are now a wide range of \emph{graph neural networks} (GNNs) for making predictions involving relational data~\citep{Battaglia2018RelationalIB,Wu2020ACS}.
These models have had much success and sit atop leaderboards such as the Open Graph Benchmark~\citep{Hu2020OpenGB}.
Often, the methodological developments for GNNs revolve around creating strictly more expressive architectures
than basic variants such as the Graph Convolutional Network (GCN)~\citep{kipf2017semi} or GraphSAGE~\citep{hamilton2017inductive};
examples include
Graph Attention Networks~\citep{velickovic2018graph},
Graph Isomorphism Networks~\citep{xu2018powerful},
and various deep models~\citep{li2019deepgcns,rong2019dropedge,chen2020simple}.
Many ideas for new GNN architectures are adapted from new architectures in models for language (e.g., attention) or vision (e.g., deep CNNs)
with the hopes that success will translate to graphs.
However, as these models become more complex, 
understanding their performance gains is a major challenge, and
scaling them to large datasets is difficult.

\begin{figure}[t]
    \centering
    \includegraphics[width=0.80\linewidth]{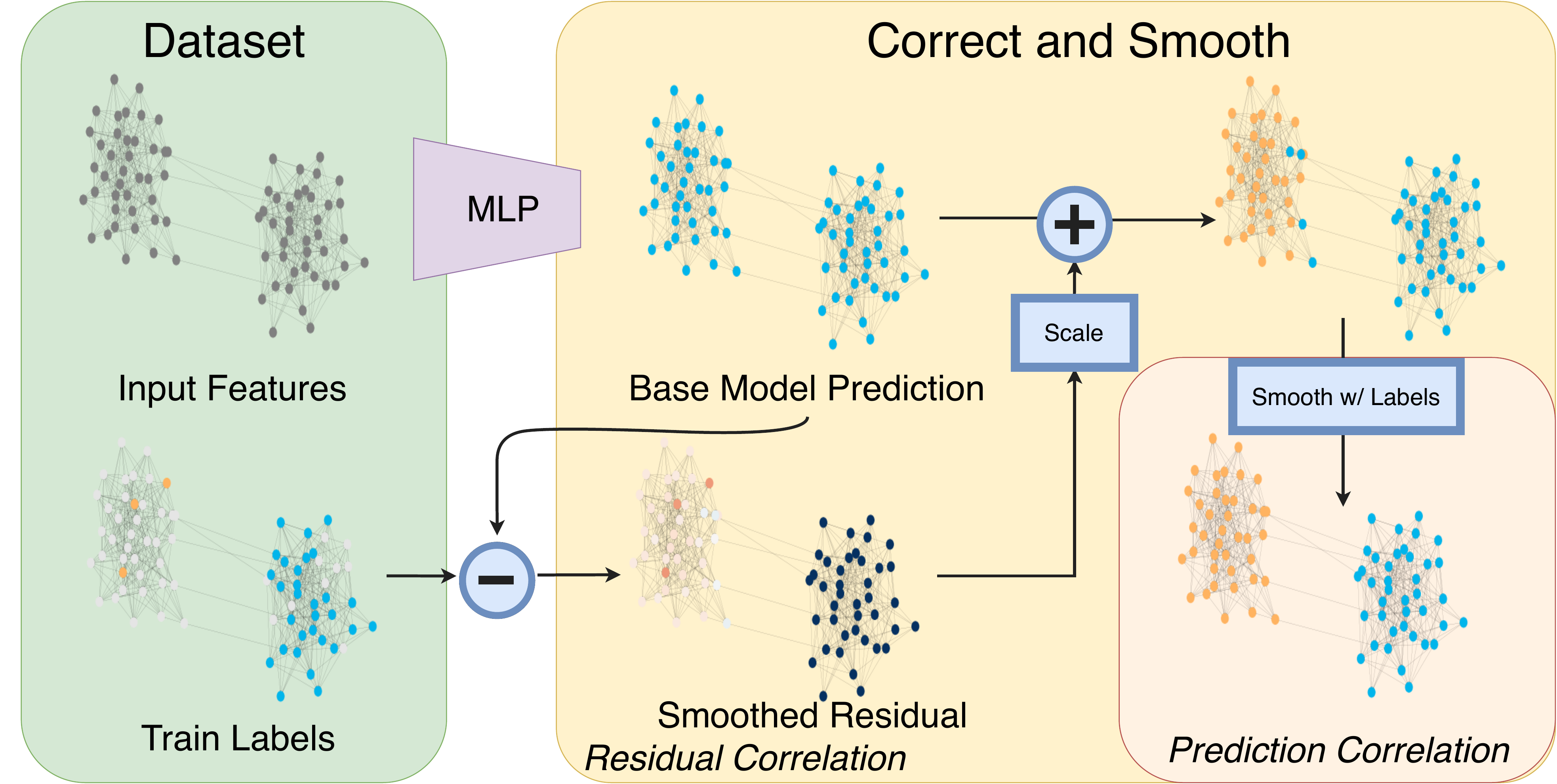}
    \caption{Overview of our GNN-free model, Correct and Smooth, with a toy example. The left cluster belongs to orange and the right cluster belongs to blue. We use MLPs for base predictions, ignoring the graph structure,
    which we assume gives the same prediction on all nodes in this example. After, base predictions are corrected by propagating errors from the training data. Finally, corrected
    predictions are smoothed with label propagation.}
    \label{fig:overview}
\end{figure}

Here, we see how far we can get by combining much simpler models,
with an emphasis on understanding where there are easy opportunities for performance improvements in graph learning, particularly transductive node classification.
We propose a simple pipeline with three main parts (\cref{fig:overview}):
(i) a base prediction made with node features that ignores the graph structure (e.g., an MLP or linear model);
(ii) a correction step, which propagates uncertainties from the training data across the graph to correct the base prediction; and
(iii) a smoothing of the predictions over the graph.
Steps (ii) and (iii) are just post-processing
and use classical methods for graph-based semi-supervised learning, namely, label propagation~\citep{zhu2005semi}.%
\footnote{One of the main methods that we use~\citep{zhou2004learning} is often called \emph{label spreading}.
The term ``label propagation'' is used in a variety of contexts~\citep{zhu2005semi,wang2007label,raghavan2007near,gleich2015using}.
The salient point for this paper is that we assume positive correlations on neighboring nodes and that the algorithms work by
``propagating'' information from one node to another.}
With a few modifications and new deployment of these classic ideas, 
we achieve state-of-the-art performance on several node classification tasks,
outperforming big GNN models. 
In our framework, the graph structure is not used to learn parameters but instead as a post-processing mechanism.
This simplicity leads to models with orders of magnitude fewer parameters that take orders of magnitude less time to train
and can easily scale to large graphs.
We can also combine our ideas with state-of-the-art GNNs and see modest performance gains.

A major source of our performance improvements is directly using labels for predictions.
This idea is not new --- early diffusion-based semi-supervised learning algorithms on graphs such as
the spectral graph transducer~\citep{joachims2003transductive},
Gaussian random field models~\citep{Zhu2003SemiSupervisedLU}, and
and label spreading~\citep{zhou2004learning}
all use this idea.
However, the motivation for these methods was semi-supervised learning on point cloud data,
so the features were used to construct the graph.
Since then, these techniques have been used for learning on relational data from just the labels (i.e., no features)~\citep{koutra2011unifying,gleich2015using,peel2017graph,chin2019decoupled}
but have largely been ignored in GNNs. 
That being said, we find that even simple label propagation (which ignores features) does surprisingly well on a number of benchmarks.
This provides motivation for combining two orthogonal sources of prediction power --- one coming from the node features (ignoring graph structure)
and one coming from using the known labels directly in predictions.

Recent research connects GNNs to
label propagation~\citep{wang2020unifying,Jia-2020-GNNR} as well as
Markov Random fields~\citep{qu2019gmnn,gao2019conditional}, and
some techniques use ad hoc incorporation of label information in the features~\citep{shi2020masked}.
However, these approaches are still expensive to train,
while we use label propagation in two understandable and low-cost ways.
We start with a cheap ``base prediction'' from a model that ignores graph structure 
(apart from perhaps a cheap pre-processing feature augmentation step like a spectral embedding).
After, we use label propagation for error correction and then to smooth final predictions.
These post-processing steps are based on the fact that errors and labels on connected nodes are positively correlated.
Assuming similarity between connected nodes is at the center of much network analysis
and corresponds to homophily or assortative mixing~\citep{mcpherson2001birds,newman2003mixing,easley2010networks}.
In the semi-supervised learning literature, the analog is the 
smoothness or cluster assumption~\citep{chapelle2003cluster,zhu2005semi}. 
The good performance of label propagation that we see across a wide variety of datasets 
suggests that these correlations hold on common benchmarks.

Overall, our methodology demonstrates that combining several simple ideas yields excellent performance in
transductive node classification at a fraction of the cost, in terms of both model size (i.e., number of parameters) and training time.
For example, on the OGB-Products benchmark, we out-perform the current best-known GNN with 
more than two orders of magnitude fewer parameters and more than two orders of magnitude less training time.
However, our goal is \emph{not} to say that current graph learning methods are poor or inappropriate.
Instead, we aim to highlight easier ways in which to improve prediction performance in graph learning
and to better understand the source of performance gains.
Our main finding is that more direct incorporation of labels into the learning algorithms is key.
And by combining our ideas with existing GNNs, we also see improvements, although they are minor.
We hope that our approach spurs new ideas that can help in other graph learning tasks, 
such as inductive node classification, link prediction, and graph prediction.

\subsection{Additional related work}

The Approximate Personalized Propagation of Neural Predictions (APPNP) framework is most
relevant to our work, as they also smooth base predictions~\citep{klicpera2018predict}.
However, they focus on integrating this smoothing into the training process so that their model can be trained end to end. 
Not only is this significantly more computationally expensive, it also prevents APPNP from incorporating label information at inference. 
Compared to APPNP, our framework produces more accurate predictions, is faster to train, and more easily scales to large datasets.
Our framework also complements the Simplified Graph Convolution~\citep{pmlr-v97-wu19e}, as well as
algorithms designed to increase scalability~\citep{bojchevski2020scaling,zeng2019graphsaint,rossi2020sign}.
The primary focus of our approach, however, is using labels directly, and scalability is a byproduct.
There is also prior work connecting GCNs and label propagation.
\Citet{wang2020unifying} use label propagation as a pre-processing step to weight edges for GNNs,
whereas we use label propagation as a post-processing step and avoid GNNs.
\Citet{Jia-2020-GNNR} use label propagation with GNNs for regression tasks, and our error
correction step adapts some of their ideas for the case of classification.
Finally, there are several recent approaches that incorporate nonlinearity into label
propagation methods to compete with GNNs and achieve
scalability~\citep{eliav2018bootstrapped,ibrahim2019nonlinear,tudisco2020nonlinear},
but these methods focus on settings of low label rates and don't incorporate feature learning.

\section{Correct and Smooth Model}\label{sec:basic}

We start with some notation.
We assume that we have an undirected graph $G = (V, E)$, where there are $n = \lvert V \rvert$ nodes
with features on each node represented by a matrix $X \in \mathbb{R}^{n \times p}$.
Let $A$ be the adjacency matrix of the graph, $D$ be the diagonal degree matrix, and
$S$ be the normalized adjacency matrix $D^{-1/2}AD^{-1/2}$.
For the prediction problem, the node set $V$ is split into a disjoint set of unlabeled nodes $U$ and labeled nodes $L$,
which are subsets of the indices $\{1, \ldots, n\}$.
We represent the labels by a one-hot-encoding matrix $Y \in \mathbb{R}^{n \times c}$, where $c$ is the number of classes
(i.e., $Y_{ij} = 1$ if $i \in L$ is in class $j$, and 0 otherwise), and
we further split the labeled nodes into a training set $L_t$ and validation set $L_v$.
Our problem is transductive node classification: 
assign each node $j \in U$ a label in $\{1, \ldots, C\}$, given $G$, $X$, and $Y$.

Our approach starts with a simple base predictor on node features, which does not rely on any learning over the graph.
After, we perform two types of label propagation (LP): one that corrects the base predictions by modeling correlated error
and one that smooths the final prediction. We call the combination of these two methods Correct and Smooth (\framework{}; \cref{fig:overview}).
The LPs are only post-processing steps --- our pipeline is \emph{not} trained end-to-end.
Furthermore, the graph is only used in these post-processing steps and in a pre-processing step to augment the features $X$, 
but not for the base predictions.
This makes training fast and scalable compared to standard GNN models.
Moreover, we take advantage of both LP (which tends to perform fairly well on its own without features)
and the node features.
We will see that combining these complementary signals yields excellent predictions.

\subsection{Simple base predictor}
To start, we use a simple base predictor that does not rely on the graph structure.
More specifically, we train a model $f$ to minimize $\sum_{i \in L_t} \ell(f(x_i), y_i)$, where $x_i$ is the $i$th row of $X$,
$y_i$ is the $i$th row of $Y$, and $\ell$ is a loss function. 
For this paper, $f$ is either a linear model or a shallow multi-layer perceptron (MLP) followed by a softmax, and $\ell$ is the cross-entropy loss.
The validation set $L_v$ is used to tune hyperparameters such as learning rates and the hidden layer dimensions for the MLP.
From $f$, we get a \emph{base prediction} $Z \in \mathbb{R}^{n \times c}$,
where each row of $Z$ is a probability distribution resulting from the softmax.
Omitting the graph structure for these base predictions avoids the scalability issues with GNNs.
In principle, though, we can use any base predictor for $Z$, including those based on GNNs,
and we explore this in \cref{sec:full}. 
However, for our pipeline to be simple and scalable,
we just use linear classifiers or MLPs with subsequent post-processing, which we describe next.

\subsection{Correcting for error in base predictions with residual propagation}
Next, we improve the accuracy of the base prediction $Z$ by incorporating labels to correlate errors.
The key idea is that we expect \emph{errors} in the base prediction to be positively correlated along edges in the graph.
In other words, an error at node $i$ increases the chance of a similar error at neighboring nodes of $i$.
We should ``spread'' such uncertainty over the graph.
Our approach here is inspired in part by residual propagation~\citep{Jia-2020-GNNR}, 
where a similar concept is used for node regression tasks, 
as well as generalized least squares and correlated error models more broadly~\citep{shalizi2013advanced}.

To this end, we first define an error matrix $E \in \mathbb{R}^{n \times c}$, 
where error is the \emph{residual} on the training data and zero elsewhere:
\begin{equation}
E_{L_t} = Z_{L_t} - Y_{L_t},\quad E_{L_v} = 0,\quad E_{U} = 0.
\end{equation}
The residuals in rows of $E$ corresponding to training nodes are zero only when the base predictor makes a perfect predictions.
We smooth the error using the label spreading technique of \citet{zhou2004learning},
optimizing the objective
\begin{equation}
\hat{E} = \argmin_{W \in \mathbb{R}^{n \times c}} \text{trace}(W^T(I - S)W) +  \mu \| W - E \|_F^2.
\end{equation}
The first term encourages smoothness of the error estimation over the graph, and is equal
to $\sum_{j=1}^{c}w_j^T(I - S)w_j$, where $w_j$ is the $j$th column of $W$.
The second term keeps the solution close to the initial guess $E$ of the error. As in \citet{zhou2004learning}, 
the solution can be obtained via the iteration $E^{(t+1)} = (1- \alpha) E  + \alpha S E^{(t)}$,
where $\alpha = 1 / (1 + \mu)$ and $E^{(0)} = E$, which converges rapidly to $\hat{E}$.
This iteration is a diffusion, propagation, or spreading of the error,
and we add the smoothed errors to the base prediction to get corrected predictions $Z^{(r)} = Z + \hat{E}$.
We emphasize that this is a post-processing technique and there is no coupled training with the base predictions.

This type of propagation is provably the right approach under a Gaussian assumption in regression problems~\citep{Jia-2020-GNNR};
however, for the classification problems we consider, the smoothed errors $\hat{E}$ might not be at the right scale. 
We know that in general,
\begin{equation}
\| E^{(t+1)} \|_2 \leq (1- \alpha) \| E \|  + \alpha \| S \|_2  \| E^{(t)}\|_2 = (1- \alpha) \| E \|_2 + \alpha \| E^{(t)}\|_2.
\end{equation}
When $E^{(0)} = E$, we then have that $\| E^{(t)} \|_2 \leq \| E \|_2$.
Thus, the propagation cannot completely correct the errors on all nodes in the graph, as it does not have enough ``total mass,''
and we find that adjusting the scale of the residual can help substantially in practice.
To do this, we propose two variations of scaling the residual.

\xhdr{Autoscale} Intuitively, we want to scale the size of errors in $\hat{E}$ to be approximately the size of the errors in $E$.
We only know the true errors at labeled nodes, so we approximate the scale with the average error over the training nodes.
Formally, let $e_j \in \mathbb{R}^c$ correspond to the $j$th row of $E$, and define
$\sigma = \frac{1}{\lvert L_t \rvert}\sum_{j \in L_t} \| e_j \|_1$. Then the corrected predictions on
an unlabeled node $i$ is given by
$Z^{(r)}_{i,:} = Z_{i,:} + \sigma \hat{E}_{:,i} / \|\hat{E}_{:,i}^T\|_1$ for $i \in U$.

\xhdr{Scaled Fixed Diffusion (FDiff-scale)}
Alternatively, we can use a diffusion like the one from \citet{Zhu2003SemiSupervisedLU},
which keeps the known errors at training nodes \emph{fixed}.
More specifically, we iterate $E^{(t+1)}_U = [D^{-1}A E^{(t)}]_U$ and keep fixed $E^{(t)}_{L} = E_{L}$ until convergence to $\hat{E}$, starting with $E^{(0)} = E$.
Intuitively, this fixes error values where we know the error (on the labeled nodes $L$), while other nodes keep averaging over the values of their neighbors until convergence.
With this type of propagation, the maximum and minimum values of entries in $E^{(t)}$ do not go beyond those in $E_L$. 
We still find it effective to learn a scaling hyperparameter $s$ to produce $Z^{(r)} = Z + s\hat{E}$. 

\subsection{Smoothing final predictions with Prediction Correlation}
At this point, we have a score vector $Z^{(r)}$, obtained from correcting the base predictor $Z$
with a model for the correlated error $\hat{E}$.
To make a final prediction, we further smooth the corrected predictions. 
The motivation is that adjacent nodes in the graph are likely to have similar labels,
which is expected given homophily or assortative properties of a network.
Thus, we can encourage smoothness over the distribution over labels by another label propagation.
First, we start with our best guess $G \in \mathbb{R}^{n \times c}$ of the labels:
\begin{equation}
G_{L_t} = Y_{L_t}, \quad G_{L_v, U} = Z^{(r)}_{L_v, U}. \label{eq:guess}
\end{equation}
Here, we set the training nodes back to their true labels and use the corrected predictions for the validation and unlabeled nodes
(we can also use the true validation labels, which we discuss later in the experiments).
We then iterate $G^{(t+1)} = (1- \alpha) G + \alpha S G^{(t)}$ with $G^{(0)} = G$ until convergence to give the final prediction $\hat{Y}$.
The classification for a node $i \in U$ is $\argmax_{j \in \{1,\ldots,c\}} \hat{Y}_{ij}$.

As with error correlation, the smoothing here is a post-processing step, decoupled from the other steps.
This type of prediction smoothing is similar in spirit to APPNP~\citep{klicpera2018predict}, which we compare against later.
However, APPNP is trained end-to-end, propagates on final-layer representations instead of softmaxes, does not use labels, and is motivated differently.

\subsection{Summary and additional considerations}
To review our pipeline, we start with a cheap base prediction $Z$, using only node features but not the graph structure.
After, we estimate errors $\hat{E}$ by propagating known errors on the training data,
resulting in error-corrected predictions $Z^{(r)} = Z + \hat{E}$.
Finally, we treat these as score vectors on unlabeled nodes, and combine them
with the known labels through another LP step to produce smoothed final predictions.
We refer to this general pipeline as \textbf{Correct and Smooth} (C\&S).

Before showing that this pipeline achieves state-of-the-art performance
on transductive node classification, we briefly describe another simple way of improving performance:
feature augmentation.
The hallmark of deep learning is that we can learn features instead of engineering them.
However, GNNs still rely on informative input features to make predictions.
There are numerous ways to get useful features from just the graph topology to augment the raw node features~\citep{henderson2011s,henderson2012rolx,Hamilton-2017-representation}.
In our pipeline, we augment features with a regularized spectral embedding~\citep{chaudhuri2012spectral,zhang2018understanding} 
coming from the leading $k$ eigenvectors of the matrix
$D_{\tau}^{-1/2}(A + \frac{\tau}{n}\mathbf{1}\mathbf{1}^T)D_{\tau}^{-1/2}$,
where $\mathbf{1}$ is a vector of all ones, 
$\tau$ is a regularization parameter set to the average degree, and
$D_{\tau}$ is diagonal with $i$th diagonal entry equal to $D_{ii} + \tau$.
The underlying matrix is dense, but we can apply matrix-vector products in time linear in the number of edges
and use iterative eigensolvers to compute the embeddings quickly.

\section{Experiments on Transductive Node Classification}\label{sec:full}
\begin{table}[t]
\caption{Summary statistics of datasets.
We include (i) the reduction in the number of parameters, 
(ii) the change in accuracy of our best \framework{} model compared to the state-of-the-art GNN method, and
(iii) the training time.
By avoiding expensive GNNs, our methods require fewer parameters and are faster to train.
Our methods are typically more accurate (see also \cref{tab:full,tab:full_all}).
}
\vspace{-\baselineskip}
\label{tab:stats}
\begin{center}
\begin{tabular}{ll ccccc}
\toprule
Datasets  & Classes & Nodes & Edges & Parameter $\Delta$ & Accuracy $\Delta$ &  Time \\
\midrule
Arxiv    & 40 & 169,343 & 1,166,243 & -84.9\% & + 0.97 & 9.89 s\\
Products   & 47 & 2,449,029 &	61,859,140 & -93.47\% & +1.53 & 170.6 s\\
Cora   & 7 & 2,708	& 5,429	& -98.37\% & + 1.09 & 0.5 s\\
Citeseer    & 6 & 3,327	& 4,732	& -89.68\% & - 0.69 & 0.48 s\\
Pubmed    & 3 & 19,717	& 44,338 & -96.00\% & - 0.30 & 0.85 s \\
Email   & 42 & 1,005 & 25,571 & - 97.89\% & + 4.26 & 42.83 s\\
Rice31   &  10 & 4,087 & 184,828& - 99.02\% & + 1.39 & 39.33 s \\
US County    & 2 & 3,234 & 12,717 & - 74.56\% & + 1.77 & 39.05 s\\
wikiCS   &  10 & 11,701 & 216,123 & - 84.88\% & + 2.03 & 7.09 s\\
\bottomrule
\end{tabular}
\end{center}
\end{table}

To demonstrate the effectiveness of our methods, we use nine datasets (\cref{tab:stats}).
The Arxiv and Products datasets are from the Open Graph Benchmark (OGB)~\citep{Hu2020OpenGB};
the Cora, Citeseer, and Pubmed are three classic citation network benchmarks~\citep{getoor2001learning,getoor2005link,namata2012query}; and
wikiCS is a web graph~\citep{Mernyei2020WikiCSAW}.
In these datasets, classes are categories of papers, products, or pages, and features are derived from text.
We also use a Facebook social network of Rice University, where classes are dorm residences
and features are attributes such as gender, major, and class year, amongst others~\citep{Traud2011SocialSO},
as well as a geographic dataset of US counties where classes are 2016 election outcomes
and features are demographic~\citep{Jia-2020-GNNR}.
Finally, we use an email dataset of a European research institute, where classes are department membership
and there are no features~\citep{leskovec2007graph,yin2017local}.

\xhdr{Data splits} The training/validation/test splits for Arxiv and Products are given by the benchmark,
and the splits for wikiCS come from \citet{Mernyei2020WikiCSAW}.
For the Rice, US counties, and email data, we use 40\%/10\%/50\% random splits,
and for the smaller citation networks, we use 60\%/20\%/20\% random splits, as in \citet{wang2020unifying} (in contrast to lower label rate settings~\citep{Yang2016RevisitingSL})
to ameliorate sensitivity to hyperparameters.
In all of our experiments, the standard deviations in prediction accuracy over splits is typically less than 1\% and does
not change our qualitative comparisons.

\xhdr{Base predictors and other models}
We use \emph{Linear} and \emph{MLP} models as simple base predictors,
where the input features are the raw node features and the spectral embedding.
We also use a \emph{Plain Linear} model that only uses the raw features for comparison
and Label Propagation (\emph{LP}; specifically, the \citet{zhou2004learning} version), which only uses labels.
For comparable GNN models to our framework (in terms of simplicity or style), we use GCN, SGC, and APPNP.
\emph{For the GCN models, we added extra residual connections from the input to every layer and from every layer to the output,
which produced better results.}
The number of layers and hidden channels for the GCNs are the same as the MLPs.
Thus, GCNs here represent a class of GCN-type models and not the original model~\cite{kipf2017semi}.

Finally, we include several ``state-of-the-art'' (SOTA) baselines.
For Arxiv and Products, this is UniMP~\citep{shi2020masked} (top of OGB leaderboard, as of October 1, 2020).
For Cora, Citeseer and Pubmed, we reuse the top performance scores from \citet{chen2020simple}.
For Email and US County, we use GCNII~\citep{chen2020simple}. 
For Rice31, we use GCN with spectral and node2vec~\citep{Grover2016node2vecSF} embeddings (this is the best GNN-based model that we found).
For wikiCS, we use APPNP as reported by \citet{Mernyei2020WikiCSAW}.
We select a set of fixed hyperparameters using the validation set.
See the appendix for additional model architecture details.

\subsection{First results on node classification}

\begin{table}[t]
\caption{Performance of our C\&S framework, using only the ground truth training labels in \cref{eq:guess}.
Further improvements can be made by including ground truth validation labels (\cref{tab:full_all}).
}\label{tab:full}
\vspace{-\baselineskip}
\begin{center}
\begin{tabular}{lcccccc}
\toprule
Methods &  Base Model  & Arxiv & Products & Cora & Citeseer &  Pubmed\\ 
  \midrule
  LP & --- & 68.5 & 74.76  & 86.50  & 70.64 & 83.74\\
 Plain GCN & --- & 71.74 & 75.64 & 85.77 & 73.68 & 88.13 \\
 SGC & Plain Linear & 69.39  & 68.83 & 86.81 & 72.04 & 84.04 \\
 APPNP & Plain Linear & 66.38 & OOM & 87.87 & 76.53 & 89.40 \\
 SOTA  & --- & \textbf{73.79} & 82.56 & 88.49 & \textbf{77.99} & \textbf{90.30} \\
\midrule
  & Plain Linear& 52.32  & 47.73  & 73.85  & 70.27 & 87.10 \\
 Base& Linear & 70.08  & 50.05  & 74.75 & 70.51 &87.19\\
 Prediction & MLP & 71.51  & 63.41  & 74.06 & 68.10 & 86.85\\
  \midrule
 & Plain Linear& 71.11  &  80.24 & 88.62 & 76.31 &  89.99 \\
 Autoscale & Linear & 72.07  & 80.25 & 88.73 & 76.75 & 89.93\\
 & MLP &  72.62	 & 78.60 & 87.39 & 76.31 & 89.33 \\
  \midrule
& Plain Linear&   70.60	& 82.54 & \textbf{89.05} & 76.22 & 89.74\\
 FDiff-  & Linear&   71.57	& 83.01 & 88.66 & 77.06& 89.51 \\
scale& MLP&   72.43	& \textbf{84.18} & 87.39 & 76.42& 89.23 \\
\midrule
Methods &  Base Model  & Email & Rice31 &  US County & wikiCS\\ 
  \midrule
  LP & --- & 70.69  & 82.19  & 87.90  & 76.72 &  \\
 Plain GCN & --- & ---  & 15.45  & 84.13  & 78.61  \\
  SGC & Plain Linear & --- & 16.59  & 83.92  & 72.86 \\
 APPNP & Plain Linear & 70.28  & 11.34   & 84.14  & 69.83  \\
 SOTA & ---  & 71.96  & 86.50  & 88.08  & \textbf{79.84} \\
\midrule
& Plain Linear& --- & 9.84  & 75.74  & 72.45  \\
 Base  & Linear & 66.24  & 70.26   & 84.07  & 74.29   \\
 Prediction & MLP & 69.13  & 17.16  & 87.70  & 73.07  \\
 \midrule
&Plain Linear &	--- &  75.99 & 85.25	&  79.57\\
 Autoscale  &Linear	& 72.50	& 86.42 & 86.15	& 79.53 \\
&MLP	& 74.55	& 85.50 & 89.64	& 78.10\\
  \midrule
 & Plain Linear & --- & 73.66 & 87.38 & 79.54 \\
 FDiff-  & Linear&   72.53 & \textbf{87.55}	& 88.11	& 79.25 \\
scale& MLP&  \textbf{75.74}	& 85.74 & \textbf{89.85} & 78.24 \\
\bottomrule
\end{tabular}
\end{center}
\end{table}

In our first set of results, we only use the training labels in our C\&S framework,
as these are what GNNs typically use to train models.
For the results discussed here, this is generous to our baselines.
The ability to include validation labels is an advantage of our approach (and label propagation in general),
and this improves performance of our framework even further (\cref{tab:stats}).
We discuss this in the next section.

\Cref{tab:full} reports the results, and we highlight a few important findings.
First, within our model, there are substantial gains from the LP post-processing steps 
(for instance, on Products, the MLP base prediction goes from 63\% to 84\%).
Second, even the Plain Linear model with C\&S is sufficient to outperform plain GCNs in many cases,
and LP (a method with no learnable parameters) is often fairly competitive with GCNs.
This is striking given that the main motivation for GCNs was to address the fact that connected
nodes may not have similar labels~\citep{kipf2017semi}.
Our results suggest that directly incorporating correlation in the graph with simple use of the features is often a better idea.
Third, our model variants can out-perform SOTA on Products, Cora, Email, Rice31, and US County (often substantially so).
On the other datasets, there is not much difference between our best-performing model and the SOTA.

To get a sense of how much using ground truth labels directly helps, we also experiment with a version of C\&S \emph{without labels}.
Instead of running our LP steps, we just smooth the output of the base predictors using the approach
of \citet{zhou2004learning} and call this the \emph{Basic Model}.
We see that the linear and MLP base predictor can often exceed the performance of a GCN (\cref{tab:basic}).
Again, these results suggest that smoothed outputs are important, and that the original motivations for GCNs are misleading.
Instead, we hypothesize that GCNs gain performance by having smoothed outputs over the graph, a similar observation
made by \citet{pmlr-v97-wu19e}.
However, there are still gaps in performance between our models here and those in \cref{tab:full} that directly use labels.
Next, we see how to improve performance of C\&S even further by using more labels.
\begin{wraptable}{r}{0.35\linewidth}
\begin{center}
\captionof{table}{Performance of our Basic Model, which only uses labels for base predictions.}
\vspace{-2.5mm}
\label{tab:basic}
\begin{tabular}{llccc}
\toprule
 Base Model   & Arxiv  &Products \\
\midrule
  Plain Linear&  63.30 & 66.27  \\
Linear & 71.42  & 78.73    \\
 MLP & \textbf{72.48} & \textbf{80.34}   \\
Plain GCN & 71.74 &  75.64 \\
\bottomrule
\end{tabular}
\end{center}
\end{wraptable}

\subsection{Further improvements by using more labels}

\begin{table}[!t]
\centering
\caption{Performance of our Correct and Smooth (\framework{}) model with both train and validation labels
ground truth labels used in \cref{eq:guess}.}\label{tab:full_all}
 \vspace{-\baselineskip}
\begin{center}
\begin{tabular}{llcccccc}
\toprule
Methods &  Base Model  & Arxiv & Products  & Cora & Citeseer & Pubmed \\ 
\midrule
& Plain Linear&	 72.71  & 80.55 & 89.54 & 76.83 & 90.01 \\
 Autoscale & Linear	& 73.78 & 80.56 & \textbf{89.77} & 77.11 & 89.98 \\
& MLP & \textbf{74.02}   & 79.29 & 88.55 & 76.36 & 89.50 \\
\midrule
& Plain Linear& 72.42   &  82.89 & 89.47 & 77.08 & 89.74 \\
 FDiff-  & Linear&  72.93   &  83.27 & 89.53 & 77.29 & 89.57\\
scale& MLP&   73.46  & \textbf{84.55} & 88.18 & 76.41 & 89.38 \\
\midrule
 SOTA  & --- & 73.65  & 82.56 & 88.49 & \textbf{77.99} & \textbf{90.30} \\
\midrule
Methods &  Base Model  & Email & Rice31 &  US County & wikiCS\\ 
\midrule
& Plain Linear&	---&  76.59 	 & 85.22 	& \textbf{81.87}\\
 Autoscale & Linear	  & 73.33  & 87.25  & 86.38  & 81.57  \\
& MLP   &  73.45  & 86.13  & 89.71 	& 80.75 \\
\midrule
& Plain Linear  &	---& 75.31 	& 88.16 	& 81.18\\
 FDiff-  &  Linear  & 72.57  & \textbf{87.89}  & 88.06 	& 81.06 \\
scale& MLP  &  \textbf{76.22}  & 86.26  & \textbf{90.05} 	& 80.83 \\
\midrule
 SOTA & --- & 71.96  & 86.50  & 88.08  & 79.84  \\
\bottomrule
\end{tabular}
\end{center}
\end{table}

We improve the \framework{} performance by using both training
and validation labels in \cref{eq:guess} instead of just the training labels. 
Importantly, we do \emph{not} use validation labels to update the base prediction model --- they are just used to select hyperparameters.
Using validation labels boosts performance even further:
\cref{tab:full_all} shows results,
\cref{tab:stats} shows gains over SOTA,
and the appendix has more details.
The ability to incorporate labels is a benefit of our approach.
On the other hand, GNNs do not have this advantage, as they 
often rely on early stopping to prevent overfitting,
may not always benefit from more data (e.g., under distributional shift),
and do not directly use labels.
Thus, our comparisons in \cref{tab:full} are more generous than needed.
With validation labels, our best model out-performs SOTA in seven of nine datasets,
often by substantial margins (\cref{tab:stats}).

The evaluation procedure for GNN benchmarks differ from those for LP.
For GNNs, a sizable validation set is often used (and needed) for substantial hyperparameter tuning, as well as early stopping.
With LP, one can use the entire set of labeled nodes $L$ with cross-validation to select the
single hyperparameter $\alpha$.
Given the setup of transductive node classification, however, there is no reason not to use validation labels at inference
if they are helpful (e.g., via LP in our case).
The results in \cref{tab:stats,tab:full_all} show the true performance of our model and is the proper point of comparison.

Overall, our results highlight two important findings.
First, big and expensive-to-train GNN models are not actually necessary for good performance for transductive node classification on many datasets.
Second, combining classical label propagation ideas with simple base predictors outperforms graph neural networks on these tasks.

\begin{figure}[!t]
\centering
\begin{minipage}{.5\linewidth}
        \centering
        \includegraphics[width=0.8\linewidth]{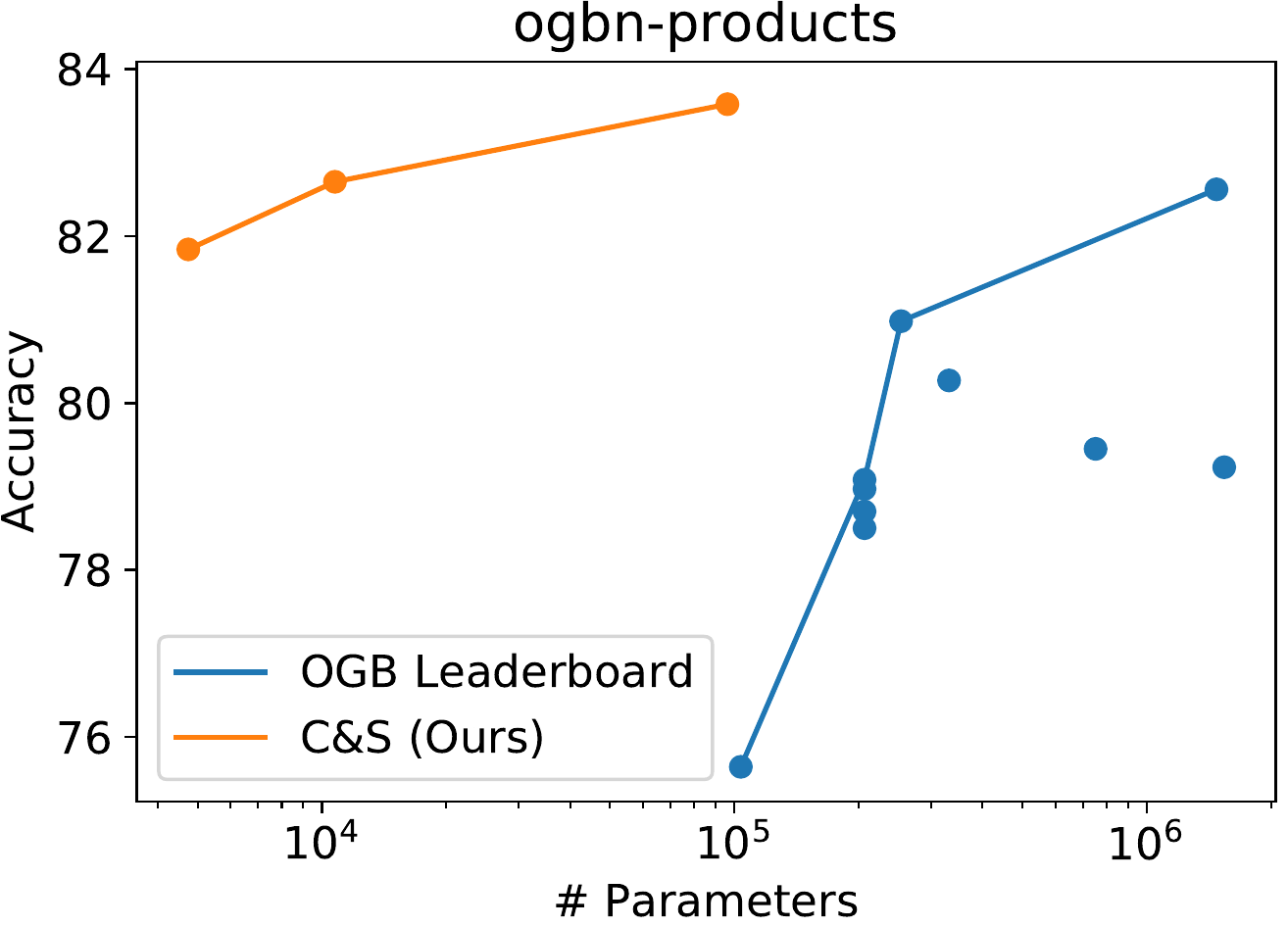}
        \label{fig:arxiv_params}
    \caption{\label{fig:parameters} Accuracy and model size on Products.} 
\end{minipage}
\hfill
\begin{minipage}{.48\linewidth}
\centering
\captionof{table}{C\&S with GNN base predictors.}\label{tab:gnn_correlation}
\vspace{-\baselineskip}
\begin{center}
\begin{tabular}{llc}
\toprule
Dataset &  Model  & Performance \\
\midrule
& GAT  & 73.56 \\
ogbn-arxiv  & GAT + \framework & \textbf{73.86} \\
 & SOTA & 73.79 \\
\midrule
US County & GCNII (SOTA) &  88.08 \\
& GCNII + \framework & \textbf{89.59} \\
\bottomrule
\end{tabular}
\end{center}
\end{minipage}
\end{figure}

\subsection{Faster training and improving existing GNNs}
\label{compute}

Our \framework{} framework often requires significantly fewer parameters compared to GNNs or other SOTA solutions. 
As an example, we plot parameters vs.\ performance for Products in \cref{fig:parameters}. 
While having fewer parameters can be useful, the real gain is in faster training time,
and our models are typically orders of magnitude faster to train than models with comparable accuracy
because we do not use the graph structure for our base predictions.
As one example, although our MLP + C\&S model based for the Arxiv dataset has a similar number of parameters compared to the GCN + labels method on the OGB leaderboards, our model runs 7 times faster per epoch and converges much faster.
In addition, compared to the SOTA for the Products dataset, 
\emph{our framework with a linear base predictor has higher accuracy, trains over 100 times faster, and has 137 times fewer parameters}. 

We also evaluated our methods on an even larger dataset, the papers100M benchmarks~\citep{Hu2020OpenGB}.
Here, we obtain 65.33\% using \framework{} with the Linear model as the base predictor, which out-performs the state-of-the-art on October 1, 2020 (63.29\%).
Due to computational limits, we could not run exhaustive benchmarks of other GNN models on this dataset.

Our pipeline can also be used to improve the performance of GNNs in general.
We applied our error correction and final prediction smoothing to more complex base predictors such as GCNII or GAT.
This improves our results on some datasets, including beating SOTA on ogbn-arxiv (\cref{tab:gnn_correlation}).
However, the performance improvements are sometimes only minor, suggesting that big models might
be capturing the same signal as our simple C\&S framework.

\subsection{Performance visualization}

\begin{figure}[t]
    \centering
    \begin{subfigure}{.5\linewidth}
      \centering
      \includegraphics[width=0.8\linewidth]{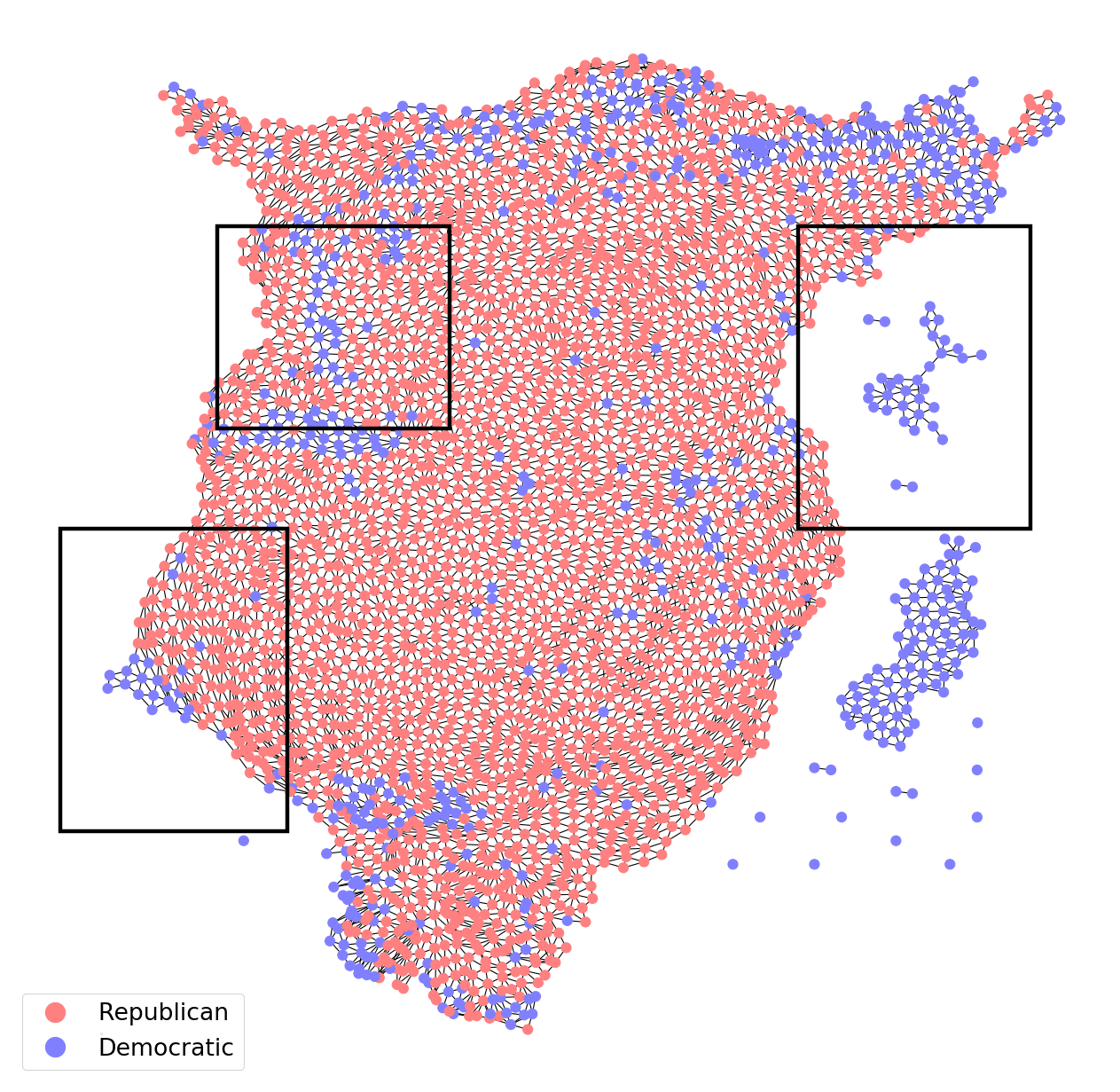}
      \caption{Ground Truth}
      \label{fig:sub1}
    \end{subfigure}%
    \begin{minipage}{.5\textwidth}
        \centering
         \begin{subfigure}{\linewidth}
        \includegraphics[width=\linewidth]{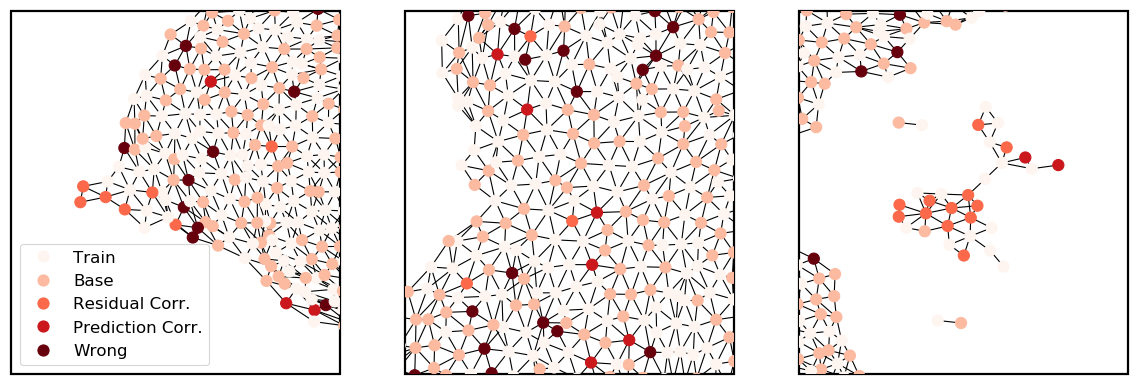}
        \caption{Spectral Embedding + Linear + \framework}
        \label{fig:sub2}
        \end{subfigure}%
        
        \begin{subfigure}{\linewidth}
        \includegraphics[width=\linewidth]{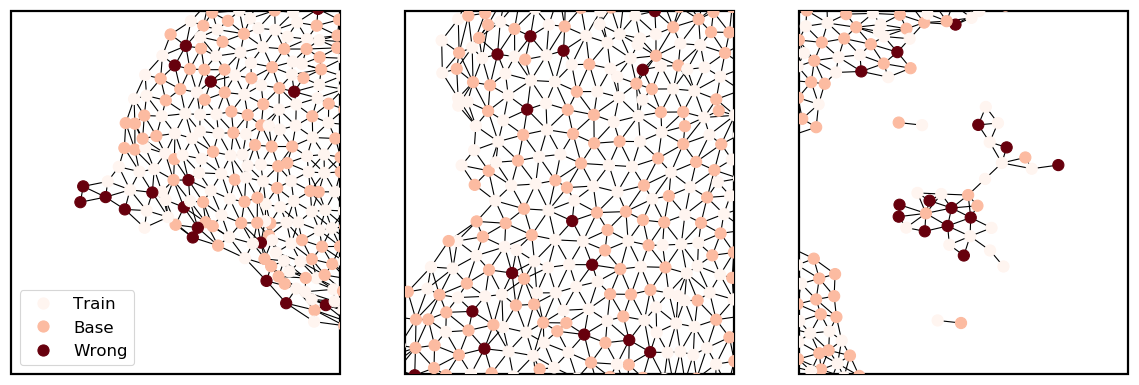}
        \caption{Spectral Embedding + GCN}
        \label{fig:sub3}
        \end{subfigure}%
    \end{minipage}%
    \caption{(a) US County visualizations, where the embedding is given by GraphViz (roughly, a compressed rotated version of the latitude and longitude coordinates). 
    Colors correspond to class labels.
    (b) Panels corresponding to parts of (a) that show at which stage \framework{} made a correct prediction.
    (c) The same panels showing GCN predictions. 
    }
    \label{fig:county_vis}
\end{figure}

To aid in understanding the performance of our \framework{} framework, we visualize the predictions on the US County dataset (\cref{fig:county_vis}).
As expected, the residual error correlation tends to correct nodes where neighboring counties provide relevant information.
For example, we see that many errors in the base predictions are corrected by the residual correlation (\cref{fig:sub2}, left and right panels)
In these cases, which correspond to parts of Texas and Hawaii,
the demographic features of the counties are outliers compared to the rest of the country, leading both the linear model and GCN astray.
The residual correlation from neighboring counties is able to fix the predictions.
We also see that the final prediction correlation will smooth the prediction, as shown in the center panel of \cref{fig:sub2}
so that the errors can be fixed based on the correct classification of the neighbors.
We observe similar behavior on the Rice31 dataset (see the appendix).

\section{Discussion}
GNN models are becoming more expressive, more parameterized, and more expensive to train.
Our results suggest that we should explore other techniques for improving performance,
such as label propagation and feature augmentation.
In particular, label propagation and its variants are longstanding, powerful ideas.
More directly incorporating them into graph learning models has major benefits,
and we have shown that these can lead to both better predictions and faster training.

\bigskip
\xhdr{Acknowledgments}
This research was supported by Facebook AI, 
NSF Award DMS-1830274, ARO Award W911NF19-1-0057, ARO MURI, and JP Morgan Chase \& Co.

We would also like to thank the rest of \href{https://cuai.github.io/}{Cornell University Artificial Intelligence} for their support and discussion. In addition, we'd like to thank Matthias Fey and Marc Brockschmidt for insightful discussions.

\bibliography{iclr2021_conference}
\bibliographystyle{iclr2021_conference}

\appendix

\section{Model Details}

Here we provide some more details on the models that we use.
In all cases we use the Adam optimizer and tune the learning rate.
We follow the models and hyperparameters provided in OGB~\citep{Hu2020OpenGB} and wikiCS~\citep{Mernyei2020WikiCSAW}
and manually tune some hyperparameters on the validation data for the potential better performance.

For our MLPs, every linear layer is followed by batch normalization, ReLU activation, and 0.5 dropout.
The other parameters depend on the dataset as follows.
\begin{itemize}
    \item OGB datasets: 3 layers and 256 hidden channels with learning rate equal to 0.01.
    
\item Cora, Citseer, Pubmed~\citep{getoor2001learning,getoor2005link,namata2012query} and Email~\citep{leskovec2007graph,yin2017local}: 3 layers and 64 hidden channels with learning rate = 0.01.

\item wikiCS: 3 layers and 256 hidden channels with learning rate equal to 0.005.

\item US County~\citep{Jia-2020-GNNR}, Rice31~\citep{Traud2011SocialSO}: 5 layers and 256 hidden channels with learning rate equal to 0.005.
\end{itemize}

Most of the ``State-of-the-Art'' models are taken from benchmark datasets.
We determined SOTA for Email, US County, and Rice31 based on all other models used in the paper.
The best performing SOTA baselines were as follows.
For Email, GCNII with 5 layers, 256 hidden channels, learning rate equal to 0.01. 
For US County, GCNII with 8 layers, 256 hidden channels, learning rate equal to 0.03. 
For Rice31, we reused our base GCN architecture and trained it over spectral and node2vec embeddings.
This significantly outperformed the other GNN variants.

All models were implemented with PyTorch\citep{pytorch} and PyTorch Geometric\citep{fey2019fast}.

\section{Performance Results with only Residual Correlation}

\Cref{tab:rc_step} shows results when using residual correlation but not smoothing in the final predictions, i.e., just the ``C'' step of our \framework{} framework.
The results indicate both the label propagation steps matter significantly for the final improvements.

\begin{table}[t]
\caption{Performance of our C\&S framework with the error correction step but not the final prediction smoothing,
using only the ground truth training labels in \cref{eq:guess}.}\label{tab:rc_step}
\vspace{-\baselineskip}
\begin{center}
\begin{tabular}{lcccccc}
\toprule
Methods &  Base Model  & Arxiv & Products & Cora & Citeseer &  Pubmed\\ 
\midrule
 & Plain Linear& 66.89 & 74.63 & 79.56 & 72.56 & 88.56 \\
 Autoscale & Linear & 71.52 & 70.93 & 79.08 & 70.77 & 88.84 \\
 & MLP & 71.97 & 69.85 & 74.11 & 71.78 & 87.35  \\
  \midrule   
& Plain Linear& 65.62 & 80.97 & 76.48 & 70.48 & 87.52 \\
FDiff-  & Linear& 70.26 & 73.89 & 79.32 & 70.53 & 84.47  \\
scale& MLP& 71.55 & 72.72 & 74.36 & 71.45 &  86.97  \\
\midrule
Methods &  Base Model  & Email & Rice31 &  US County & wikiCS\\ 
\midrule
&Plain Linear &	--- & 43.97 & 82.60 & 77.49\\
 Autoscale  &Linear	& 73.39 & 86.19 & 84.08 & 74.06\\
&MLP	& 71.64 & 84.61 & 88.83 & 78.72\\
  \midrule
 & Plain Linear & --- & 72.44 & 87.16 & 75.98 \\
 FDiff-  & Linear& 71.31 & 85.22 & 88.27  & 73.86 \\
scale& MLP& 72.59 & 85.42 & 89.62 & 78.40 \\
\bottomrule
\end{tabular}
\end{center}
\end{table}

\section{Additional visualization}

\begin{figure}[b]
      \centering
      \includegraphics[width=\linewidth]{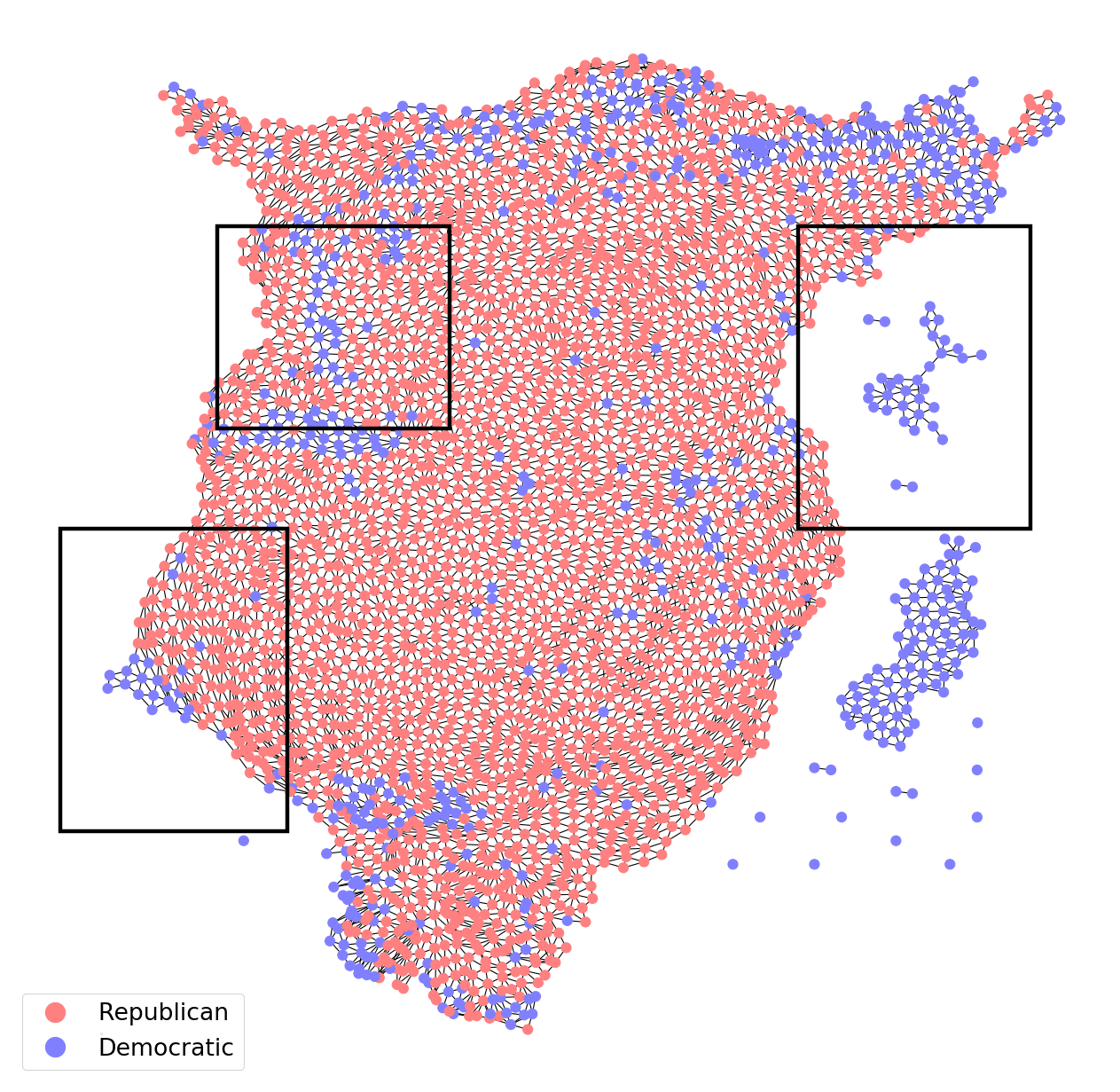}
      \caption{US County ground truth class labels.}
      \label{fig:1}
\end{figure}

\begin{figure}
      \centering
      \includegraphics[width=\linewidth]{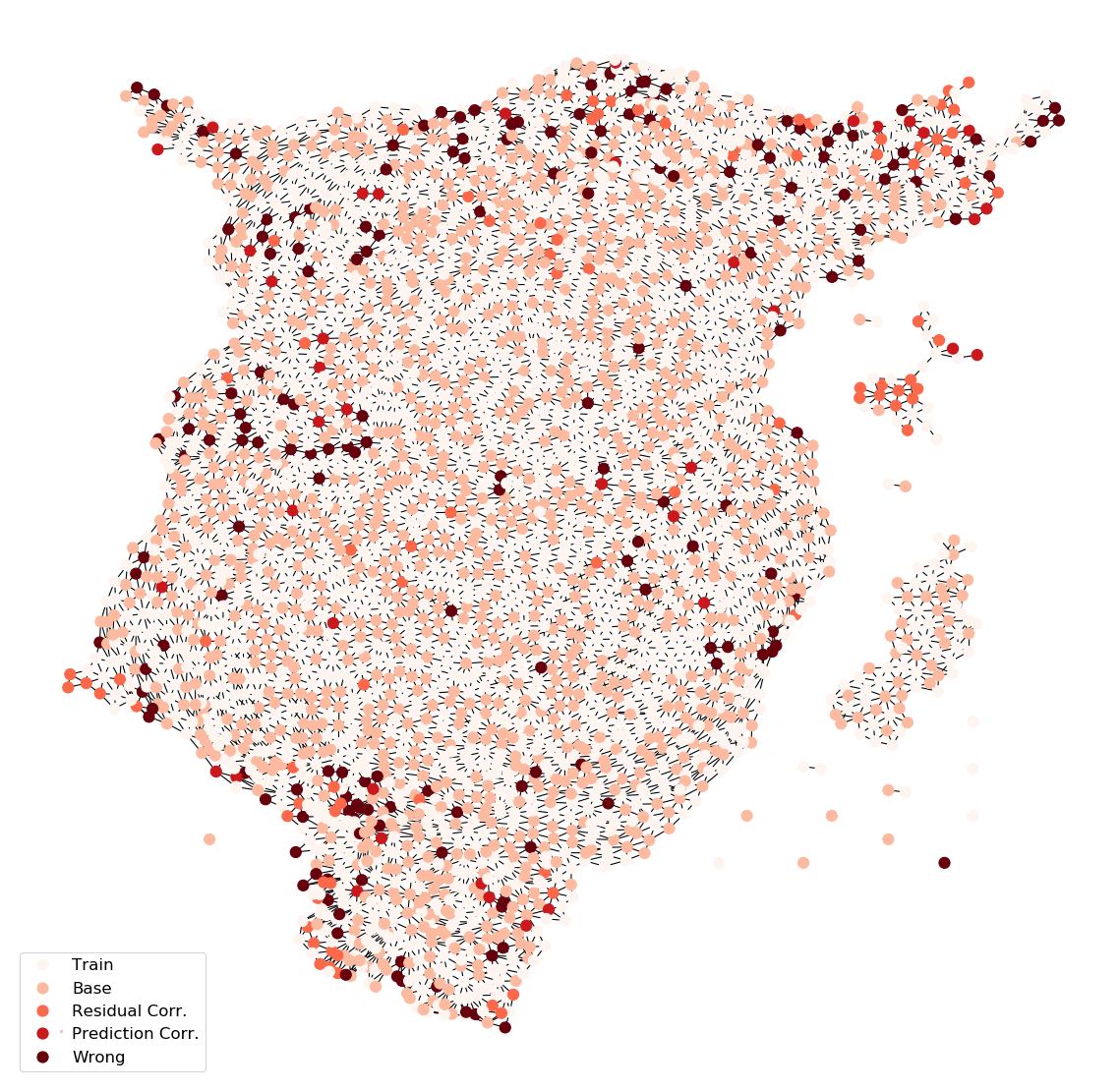}
      \caption{US County Linear Base Prediction within \framework{}.}
      \label{fig:2}
\end{figure}

\begin{figure}
      \centering
      \includegraphics[width=\linewidth]{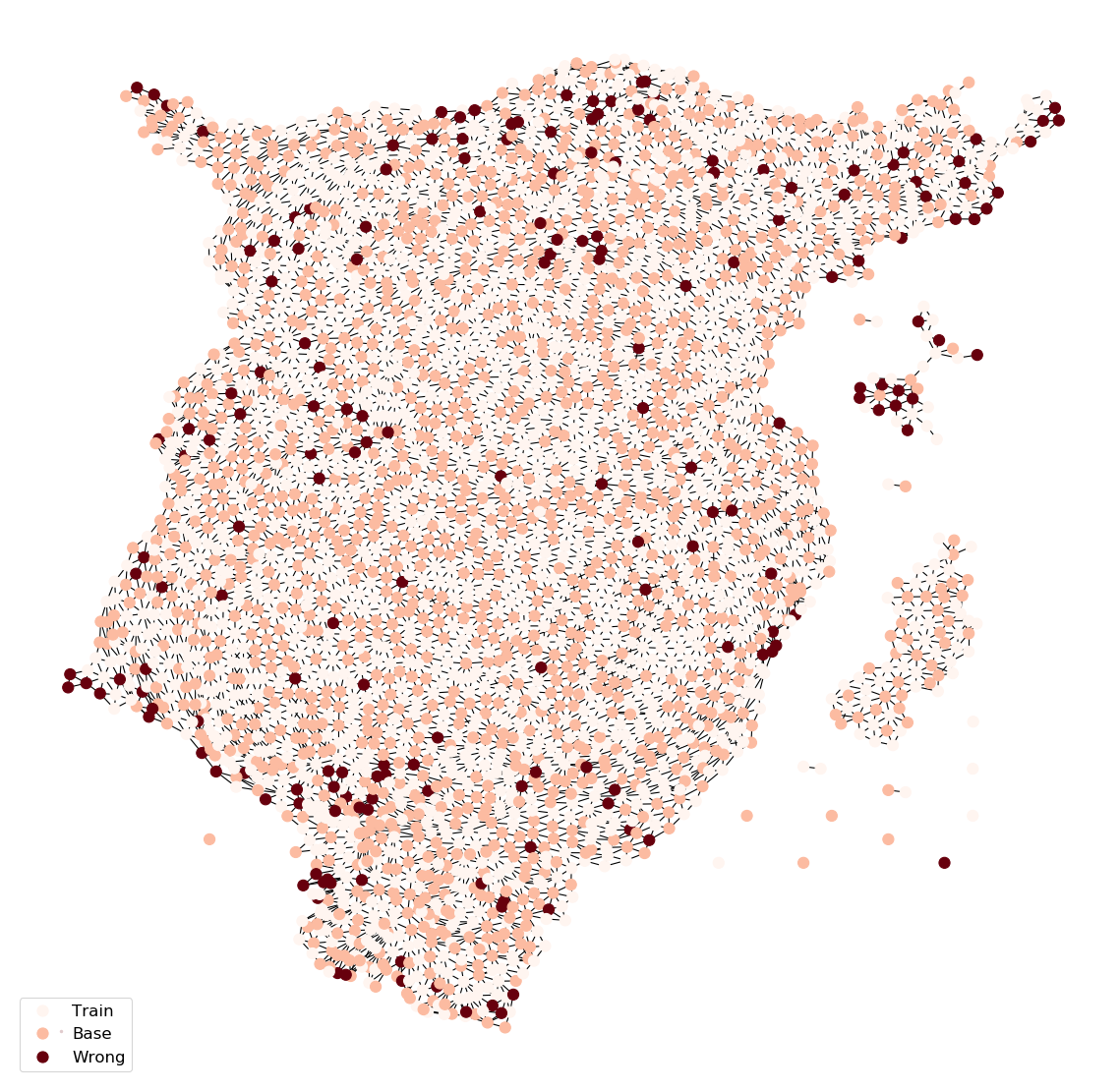}
      \caption{US County GCN (includes spectral embedding features).}
      \label{fig:3}
\end{figure}

\begin{figure}[b]
      \centering
      \includegraphics[width=\linewidth]{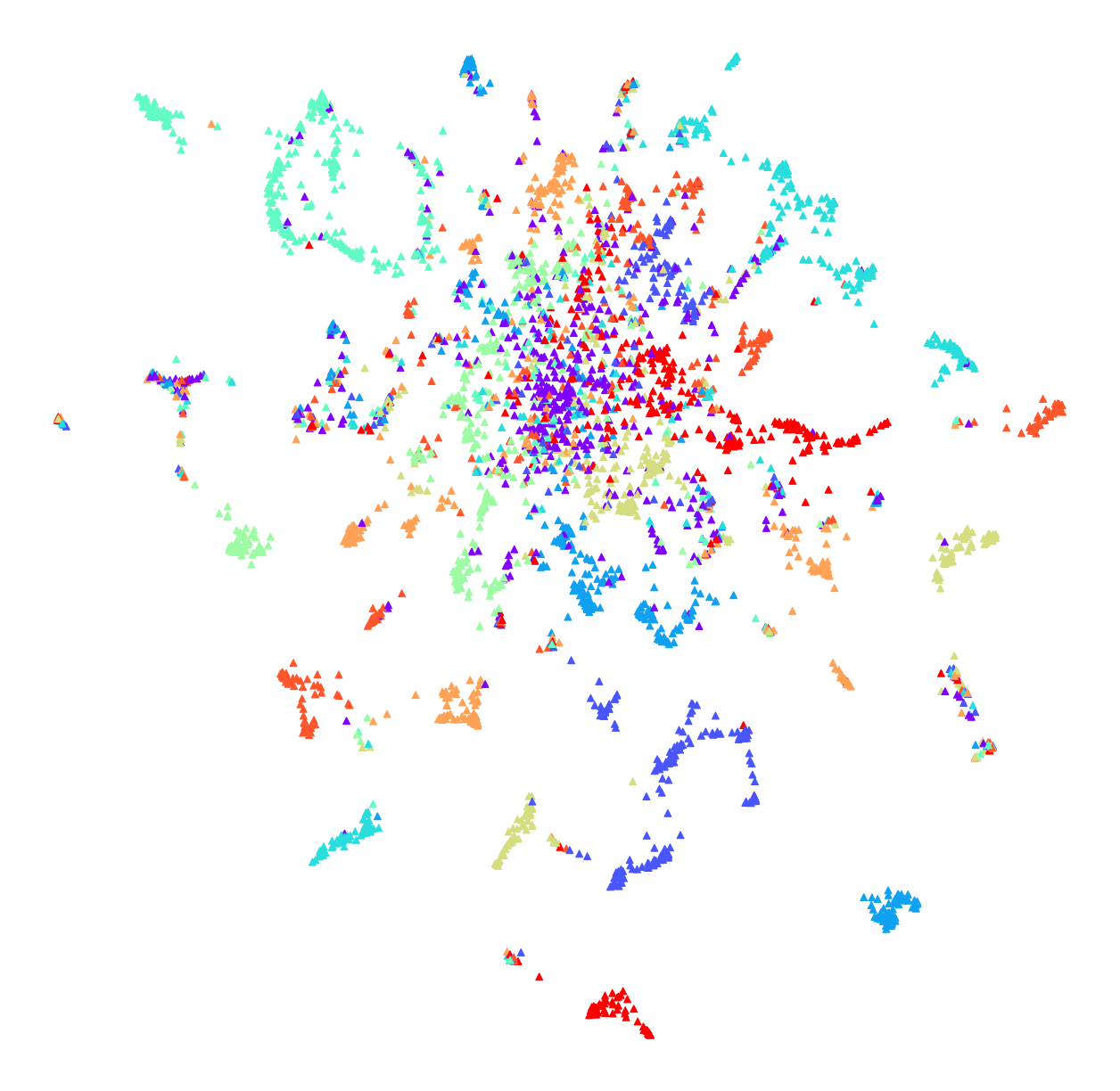}
      \caption{Rice31 ground truth class labels.}
      \label{fig:4}
\end{figure}

\begin{figure}
      \centering
      \includegraphics[width=\linewidth]{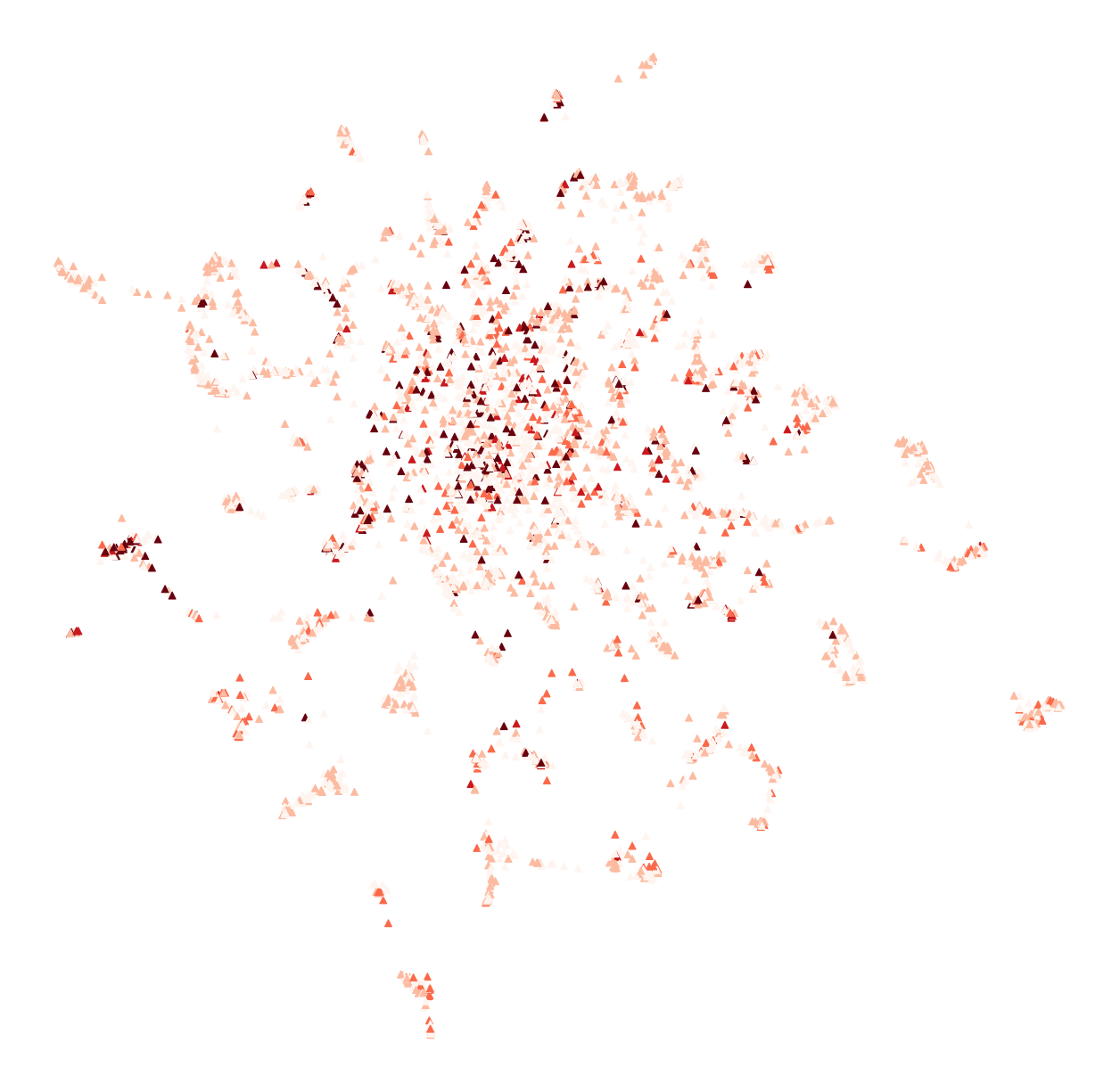}
      \caption{Rice31 Linear Base Predictor within \framework{}.}
      \label{fig:5}
\end{figure}

\begin{figure}
      \centering
      \includegraphics[width=\linewidth]{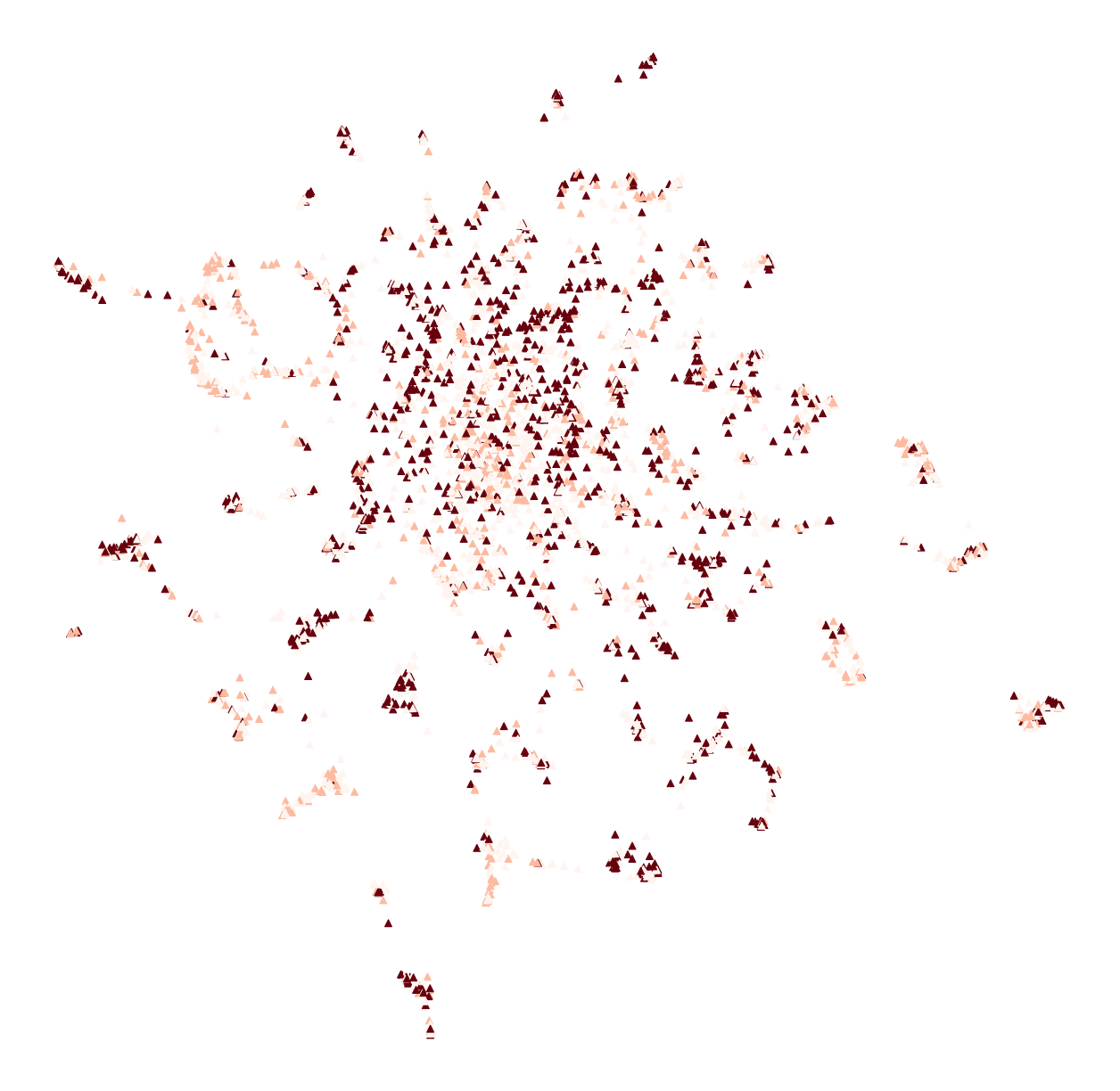}
      \caption{Rice31 GCN (includes spectral embedding features).}
      \label{fig:6}
\end{figure}

Visualizations of the US County and Rice31 dataset are shown in \cref{fig:1,fig:2,fig:3,fig:4,fig:5,fig:6}. 
The Rice31 visualization is generated by projecting the 128-dimensional spectral embedding used in the main text
down to two dimensions with UMAP~\citep{mcinnes2018umap}.

\end{document}